\documentclass[11pt]{article}

\usepackage{acl}

\usepackage{times}
\usepackage{latexsym}
\usepackage{amsmath}
\usepackage{amssymb}
\usepackage{algorithm}
\usepackage{algpseudocode}
\usepackage{booktabs}
\usepackage{array}
\usepackage{multirow}
\usepackage[table]{xcolor} 
\usepackage{nicematrix} 
\usepackage{fontawesome5} 
\usepackage{makecell}
\usepackage{stfloats}
\usepackage{enumitem}
\usepackage{cleveref}
\usepackage{subcaption} 
\usepackage{placeins}
\usepackage{microtype}

\usepackage[most]{tcolorbox}
\newtcolorbox{promptbox}[1]{
  colback=black!5,          
  colframe=black!75,        
  fonttitle=\bfseries,      
  colbacktitle=black!75,    
  coltitle=white,           
  title=#1,                 
  arc=2mm,                  
  boxrule=1pt,              
  drop shadow,              
}

\usepackage[T1]{fontenc}

\usepackage[utf8]{inputenc}

\usepackage{inconsolata}

\usepackage{graphicx}
\graphicspath{{./}{latex/}{figures/}{latex/figures/}}

\title{LLP: LLM-Based Product Pricing in E-commerce}

\author{
  \textbf{Hairu Wang\textsuperscript{1,*}}
  \quad \textbf{Sheng You\textsuperscript{1,*}}
  \quad \textbf{Qingheng Zhang\textsuperscript{1}}
  \quad \textbf{Xike Xie\textsuperscript{2}}\\
  \textbf{Shuguang Han\textsuperscript{1,$\dagger$}}
  \quad \textbf{Yuchen Wu\textsuperscript{1}}
  \quad \textbf{Fei Huang\textsuperscript{1}}
  \quad \textbf{Jufeng Chen\textsuperscript{1}}\\[2pt]
  \textsuperscript{1}Xianyu of Alibaba, Hangzhou, China\\
  \textsuperscript{2}University of Science and Technology of China, Suzhou, China\\
  \texttt{\{wanghairu.whr,yousheng.ys,qingheng.zqh,shuguang.sh\}@taobao.com}\\
  \texttt{\{wuyuchen.wyc,huangfei.hf,jufeng.cjf\}@taobao.com}\quad
  \texttt{xkxie@ustc.edu.cn}
}

\begin{document}
\maketitle
\begingroup
\renewcommand{\thefootnote}{\fnsymbol{footnote}}
\footnotetext[1]{Equal contribution.}
\footnotetext[2]{Corresponding author.}
\endgroup
\begin{abstract}
Setting appropriate prices for second-hand products remains challenging for inexperienced individual sellers on Consumer-to-Consumer (C2C) e-commerce platforms. Existing regression-based approaches rely on static product–price mappings, making it difficult to capture subtle condition differences expressed in colloquial product descriptions.
Inspired by recent breakthroughs in Large Language Models (LLMs), we present LLP, a \textit{retrieval-then-generation} framework that generates price suggestions based on dynamically retrieved comparable products. 
LLP employs bidirectional reasoning to generate high-quality rationales as supervision, and subsequently applies retrieval-aware policy optimization to further improve both reasoning faithfulness and pricing accuracy.
At inference time, LLP employs entropy-guided filtering to discard low-confidence price predictions.
Extensive evaluations demonstrate the effectiveness of our method. It raises the static adoption rate (SAR) from 40\% to 72\%, while increasing GMV by 0.7\% and CTCVR by 1.4\%.
LLP is now fully deployed in Xianyu\footnote{Xianyu is China’s largest C2C e-commerce platform.}, serving millions of individual
sellers daily.

\end{abstract}

\section{Introduction}
\begin{figure}[htbp]
    \centering
    \includegraphics[width=0.46\textwidth]{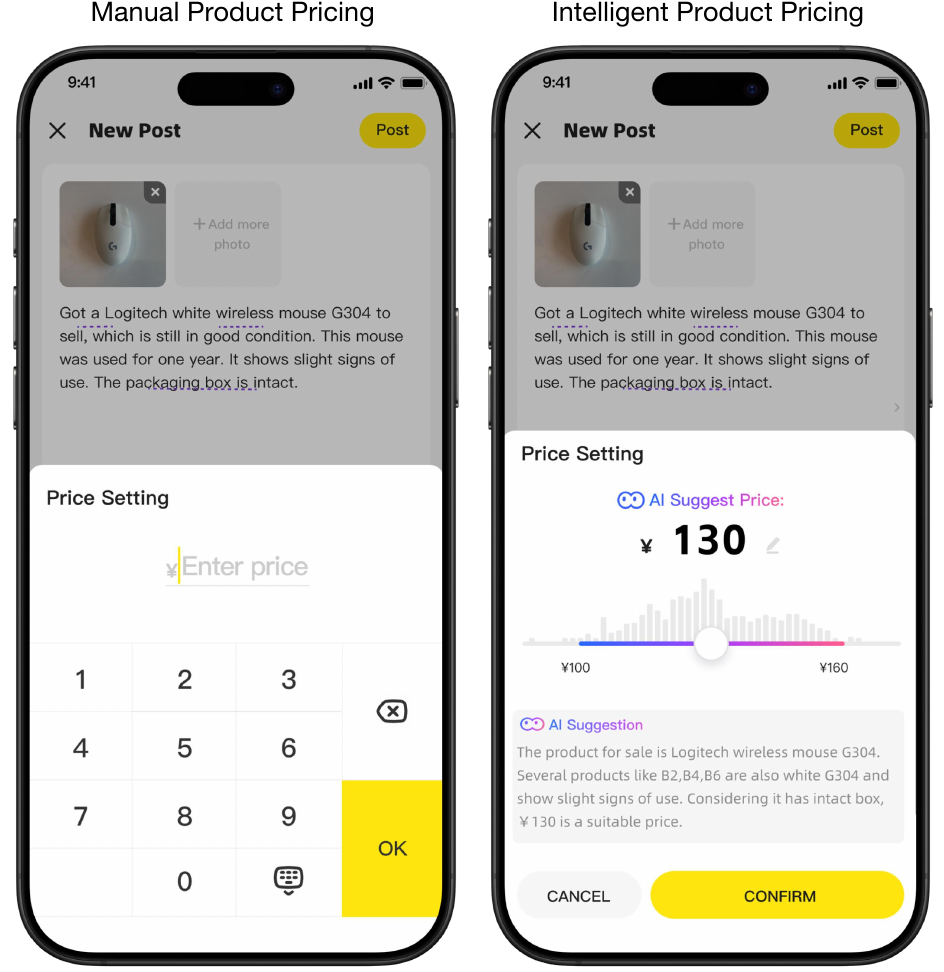}
    \caption{Intelligent Product Pricing on C2C Platforms.}
    \label{fig:llp}
\end{figure}

Driven by the sharing economy and sustainable consumption, C2C e-commerce platforms (e.g., eBay, Xianyu and Mercari) have flourished globally in recent years \cite{guiot2010second, moriuchi2022role, chen-etal-2024-ipl, 10.1145/3711896.3737237}.
Whereas business-to-consumer (B2C) platforms (e.g., Amazon) maintain deep inventories of standardized new products, C2C marketplaces aggregate highly heterogeneous second-hand items from individual sellers.
Consequently, due to the pronounced disparities in product condition and limited market knowledge, these sellers often struggle to set appropriate prices for their products \cite{raykhel2009real}.
This not only affects product listing efficiency, but also influences online transactions.

To this end, it is imperative to assist individual sellers in setting reasonable prices for their second-hand items.
If we could automatically generate price suggestions by leveraging similar products from C2C platforms, it would significantly streamline the listing process.
As shown in Figure \ref{fig:llp}, after uploading product photos and composing content descriptions, individual sellers can adopt the price with a single click or adjust it as needed, effectively eliminating the burden of manual price entry.

Leading C2C e-commerce platforms are increasingly exploring automated pricing methods. Early works \cite{raykhel2009real, 8022758, 9740817} employ traditional machine learning algorithms, such as K-Nearest Neighbors (KNN) and decision trees, to perform category-specific price prediction.
Subsequent studies \cite{7937942, 10.1145/3342240, han2019vision, han2020pricesuggestiononlinesecondhand, vedula2025quantileregressionlargelanguage} shift towards deep learning models to learn the mapping between product representations and prices, progressively extending their scope to more categories.
In our previous industrial deployment, we utilized a Category Property Value (CPV)-based pricing method~\cite{sabeh2024empirical, su2025taclrscalableefficientretrievalbased, shinzato-etal-2023-unified}, which maps target products to pre-computed clusters based on CPV attributes, followed by a Gaussian Mixture Model (GMM) for price estimation within the matched cluster.
Although effective in certain tasks, existing methods still fall short in C2C transactions, due to two primary constraints:

\textbf{Difficulty in Capturing Fine-Grained Semantics:} Unlike standardized B2C products, second-hand items exhibit high heterogeneity (e.g., varying degrees of wear and tear), which lies in colloquial product descriptions. Conventional regression models, lacking rich text understanding, struggle to capture these subtle yet critical price determinants (e.g., 85\% battery health), thereby compromising pricing accuracy.

\textbf{Limited Adaptability and Generalization:} Prices in the second-hand market are highly sensitive to product iterations and evolving market sentiment (e.g., the rapid depreciation of the iPhone 16 following the release of the iPhone 17). Furthermore, product attributes diverge significantly across categories (e.g., battery health for phones vs. remaining volume for perfumes). However, existing regression models, relying on static parameters and unified feature spaces, fail to capture such temporal or category-specific dynamics, resulting in poor adaptability and generalization.

These challenges raise a critical question: 
\textit{Can we design a system that captures fine-grained semantics and adapts to market dynamics for accurate product pricing?}

C2C platforms host a vast and diverse range of second-hand products. The Law of One Price \cite{FROOT19951647, b0277d37-409b-3c4c-9930-5213f64a850d, 0c00c0aa-2206-3f03-8569-70727b9ac7ad} in economics posits that homogeneous goods trade at a uniform price in efficient markets, implying that items of similar condition should command comparable prices. 
Moreover, autoregressive LLMs
have demonstrated remarkable success across diverse domains \cite{gu2023llm, laban-etal-2023-summedits, yi2025surveyrecentadvancesllmbased}. 
Therefore, we can adopt a "retrieval-then-generation" paradigm \cite{wang2025pathpoolingtrainingfreestructure, asai2024self, yan2024correctiveretrievalaugmentedgeneration}: retrieving similar products as dynamic market references and leveraging the nuanced semantic understanding of unstructured text to generate accurate price suggestions.
Inspired by the efficacy of Group Relative Policy Optimization (GRPO) \cite{deepseekai2025deepseekr1incentivizingreasoningcapability, hou2025t1advancinglanguagemodel, jiang2025s3dontneeddata, li2025r3raglearningstepbystepreasoning, huang2025ragrladvancingretrievalaugmentedgeneration} on verifiable tasks, we can further align the LLMs' generation with retrieved candidates, guiding them to produce more reliable, evidence-grounded price suggestions. 

Specifically, our contributions are fourfold:
\begin{itemize}[leftmargin=*, nosep]
    \item We introduce LLP for second-hand product pricing using a retrieval-then-generation paradigm, which integrates similar product retrieval with LLM-based price estimation.
            \item We propose \textit{bidirectional reasoning} to automatically construct high-quality rationale supervision. Building on this, \textit{retrieval-aware policy optimization} encourages the model to leverage retrieved products for more accurate price estimation.
    \item During inference, we develop an \textit{entropy-guided filtering} mechanism that effectively prunes low-confidence price suggestions, enabling a flexible precision--coverage trade-off in the practical application.
    \item Extensive offline and online evaluations demonstrate the effectiveness of our method. LLP has been fully deployed in production, achieving a substantial increase in the static adoption rate (SAR) from 40\% to 72\%.
\end{itemize}

\section{Preliminaries}
The abundance of listings on C2C marketplaces provides valuable references for estimating the price of a target product.
Formally, let $\mathcal{M}$ denote an LLM and $b_q$ be the target product awaiting pricing. The knowledge base is defined as $\mathcal{B}=\{(b_i,p_i)\}_{i=1}^N$, containing $N$ product-price pairs, where $b_i$ represents the details of a product (including images, textual description, and condition) and $p_i$ is its corresponding price. Our objective is to find relevant market references $\mathcal{B}_q$ from $\mathcal{B}$ by a retrieval algorithm $\mathcal{G}$: $\mathcal{B}_q=\mathcal{G}(b_q,\mathcal{B})$.
Then, a pricing algorithm $\mathcal{F}$ derives a rationale $\mathcal{R}$ and the final price estimate $\hat{p_q}$. Given the powerful language comprehension capabilities of LLMs, we instantiate $\mathcal{F}$ with $\mathcal{M}$ here: $\mathcal{R},\hat{p_q}=\mathcal{F}(b_q,\mathcal{B}_q)$.

To further enhance the reasoning capabilities of off-the-shelf LLMs for pricing tasks, we conduct domain-adaptive post-training. This process, denoted by $\mathcal{T}$, aims to produce an optimized model $\mathcal{M'}$ whose error $\mathcal{L}$ is lower than that of $\mathcal{M}$.

\begin{equation}
\mathcal{M'}=\mathcal{T}(\mathcal{M},\mathcal{B}), s.t., \mathcal{L}_{\mathcal{M'}}<\mathcal{L}_{\mathcal{M}}
\end{equation}

Moreover, we introduce an entropy-based filtering mechanism that retains a price prediction only when its average token entropy $\bar{\mathcal H}$ does not exceed a predefined threshold $\theta_{\mathcal H}$:
\begin{equation}
\hat{p}_q = \hat{p}_q \cdot \mathbb{I}(\bar{\mathcal{H}} \le \theta_{\mathcal{H}})
\end{equation}

\section{Methodology}

\begin{figure*}[htbp]
  \centering
  \includegraphics[width=\textwidth]{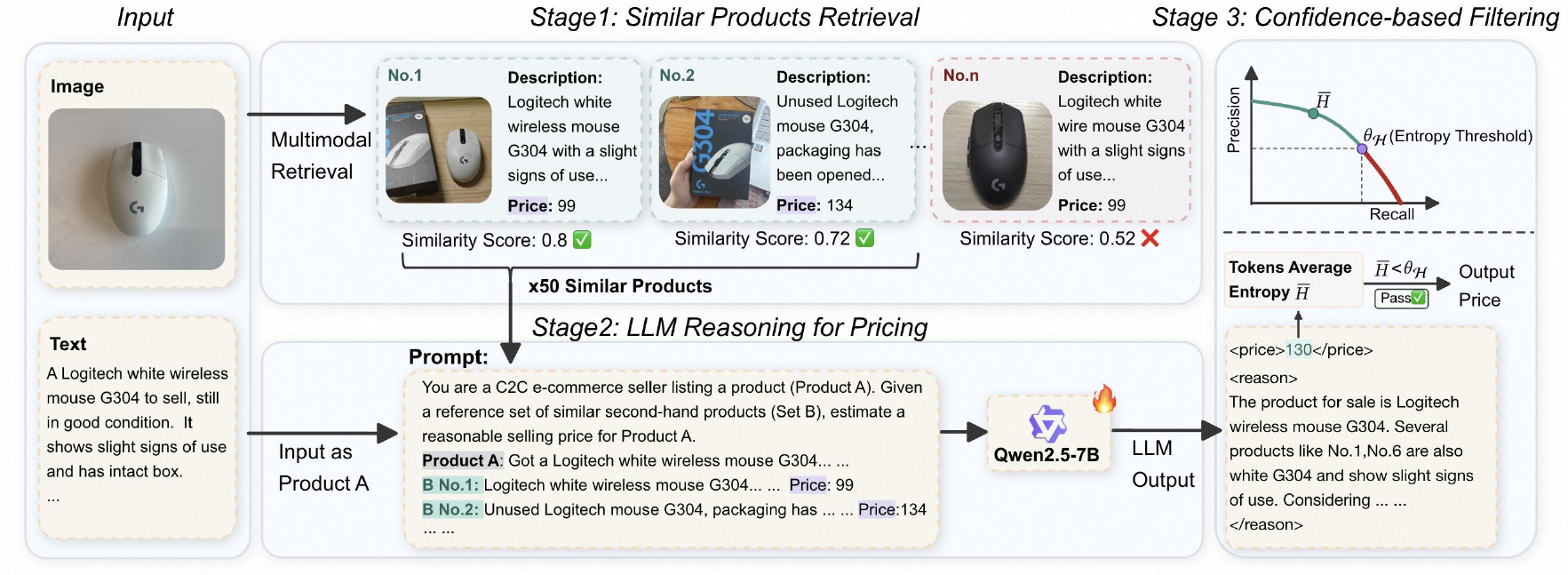} 
  \caption{LLP: LLM-based Pricing Framework for C2C E-commerce Platforms.}
  \label{fig:wide-image} 
\end{figure*}

\subsection{Overview}
In this section, we introduce the framework of our intelligent pricing system LLP, as illustrated in Figure \ref{fig:wide-image}. LLP comprises two primary components: similar product retrieval (Section \ref{retrieval}) and LLM-driven price estimation (Section \ref{reasoning}).

\subsection{Similar Product Retrieval}
\label{retrieval}
\textbf{Candidate Product Pool Construction.}\label{pool}
To provide timely and reliable market references, we construct a product pool from historical listings. First, we select products listed within the $t$-day window preceding the target product's listing date. The window is set to 90 days to preserve market recency while retaining sufficient candidates for subsequent retrieval. Beyond pool size, candidate quality is also important, as unreliable listings can degrade both retrieval relevance and downstream pricing accuracy. We therefore exclude anomalous listings, including off-platform trades, clickbait, and counterfeits. To further ensure that the candidate products reflect genuine market interest, we retain only high-engagement items whose click-through rates exceed the $c$-th percentile (e.g., $c=70$), yielding the final product pool $\mathcal{B}$.

\paragraph{Offline Product Representation Extraction.}
\label{embedding}
To efficiently retrieve similar products from $\mathcal{B}$ as dynamic market references, we employ Semantic ID \cite{yang2025gsidgenerativesemanticindexing, 10.1145/3705328.3748123, 10.1145/3701716.3715228} for representation extraction. This approach leverages a coarse-to-fine hierarchical structure that facilitates knowledge transfer and ensures robust generalization. Concretely, a T5-based encoder--decoder autoregressively generates a sequence of discrete codes via step-specific codebooks, and the decoder hidden state at the final step is taken as the product representation (see Appendix \ref{sec:semantic} for details).
For all products in the candidate product pool $\mathcal{B}=\{(b_i,p_i)\}_{i=1}^N$, we compute their representations offline and build an index repository $\mathbb{E}=\{e_i\}_{i=1}^N=\{\mathcal{E}(b_i)\}_{i=1}^N$.

\paragraph{Online Product Retrieval.}
For a newly listed product $b_q$, we map it into the same representation space to obtain the query embedding $e_q=\mathcal{E}(b_q)$.
The semantic relevance between $b_q$ and a candidate $b_i$ is quantified by cosine similarity: $\text{sim}(b_q, b_i) = \frac{\mathcal{E}(b_q)^\top \mathcal{E}(b_i)}{\|\mathcal{E}(b_q)\| \|\mathcal{E}(b_i)\|}$. To enable efficient vector search, we employ Proxima~\cite{proxima2021}, an approximate nearest-neighbor search (ANNS) engine.
Finally, for the target product \textbf{$b_q$}, the system returns the top-$k$ relevant second-hand products $\mathcal{B}_q=\{{(b_{j},p_{j})}\}_{j=1}^k$ as market references for the subsequent reasoning.
\subsection{LLM-Driven Price Inference}
\label{reasoning}
\paragraph{Prompt Engineering.}
\label{prompt}
Since retrieved products may differ from the target in key attributes, we leverage the LLM's strong semantic understanding capabilities to discern fine-grained differences and generate a well-informed price suggestion.
Our prompt incorporates three essential parts:
(1) definition and criteria of similar products, 
(2) target product description, and
(3) retrieved product descriptions and prices. The prompt templates are provided in Appendix~\ref{appendix_prompt}.
 
\paragraph{Bidirectional Reasoning-based Fine-Tuning.}
\label{sft}
To construct training data for supervised fine-tuning, we require both a price label and a corresponding pricing rationale for each target product. Specifically, we treat the seller-provided listing price $p_q$ as ground truth. To obtain the rationale $\mathcal R$, a straightforward approach is to prompt a teacher model to generate it based on the retrieved products. However, the generated $\mathcal R$ may refer to inappropriate products or fail to support the ground-truth price, resulting in inconsistencies among the retrieved products, reasoning process and price label. To construct a more coherent and evidence-grounded rationale as supervision, we introduce \textit{Bidirectional Reasoning}, a two-stage pipeline consisting of backward comparable-product selection and forward rationale synthesis.

\noindent\textit{Backward comparable-product selection.}
Specifically, in the backward stage, the teacher LLM uses the target product–price pair $(b_q, p_q)$ to select products from the retrieved set $\mathcal{B}_q$ that match $b_q$ on key attributes (e.g., brand, model, and condition) and have prices close to $p_q$. These products constitute a refined subset $\mathcal{B}_q^* \subseteq \mathcal{B}_q$.
We filter samples for which the backward reasoning returns an empty subset. Such cases typically arise when the target item's description is incomplete, too few similar products are retrieved, or the retrieved prices vary substantially, making it difficult to infer the target item's price from the retrieved evidence. \begin{equation}
    \mathcal{B}^{*}_q = \mathcal{M}_{\text{backward}}(\mathcal{B}_q, b_q, p_q)
\end{equation}

\noindent\textit{Forward rationale synthesis.}
In the subsequent forward stage, the teacher synthesizes an evidence-grounded rationale $\mathcal R$ based on the target product $b_q$ and the retrieved set $\mathcal{B}_q$, drawing primarily on the comparable products in $\mathcal{B}_q^*$. The ground-truth price $p_q$ serves only as a reference to ensure that $\mathcal R$ remains consistent with that price. To prevent label leakage and post hoc rationalization, we instruct the teacher not to use $p_q$ as a given premise or reason backward from it. Instead, every claim in $\mathcal R$ must be supported by evidence from the target product or the retrieved set. The resulting training data helps the student model learn to analyze product information and derive accurate price estimates.

\begin{equation}
    \mathcal{R} = \mathcal{M}_{\text{forward}}(\mathcal{B}_q, \mathcal{B}^*_q, b_q, p_q)
\end{equation}

To enhance second-hand e-commerce domain knowledge and retrieval-grounded reasoning capability, we fine-tune~\cite{10.1145/3485447.3511998, wang2020kadapterinfusingknowledgepretrained, zhang2024raft, clemedtson2025graphraftretrievalaugmentedfinetuning} a small student model on the constructed dataset. The generated rationale $\mathcal R$ and the target product's listing price $p_q$ are used as supervision for standard next-token prediction:
\begin{equation}
    \mathcal{L}_{\mathrm{SFT}}(\theta) = - \sum_{j=1}^{T} \log P(y_j \mid X, y_{<j}; \theta)
\end{equation}
where $X=(b_q,\mathcal{B}_q)$ denotes the input, $y=(\mathcal{R},p_q)$ represents the output sequence, and $T$ is its length.

\paragraph{Retrieval-aware Policy Optimization.}
\label{grpo}
Building on the supervised fine-tuned model, we apply GRPO to further enhance the model's reasoning capability and pricing accuracy.
In the design of the reward function, we evaluate both the generation process and the final outcome. The \textit{outcome-based reward} assigns higher scores to more accurate price estimates, decreasing as the relative prediction error increases.
However, optimizing for price accuracy alone is insufficient, as the LLM may overlook similar products or hallucinate unsupported details, resulting in unfaithful rationales.
To mitigate this issue, we introduce a \textit{process-based reward} that measures how well the products referenced in the rationale $\hat{\mathcal{B}}_q^*$ cover the comparable subset $\mathcal{B}_q^*$ selected during backward reasoning.
\begin{equation}
\begin{aligned}
reward = \underbrace{
  \frac{1}{1+\alpha\cdot\left(\frac{\hat{p}_q-p_q}{p_q}\right)^2}
  }_{\text{price accuracy}}
  \times
  \underbrace{
  \frac{
\left|\widehat{\mathcal{B}}^{*}_q\cap\mathcal{B}^{*}_q\right|
  }{
  \left|\mathcal{B}^{*}_q\right|
  }
  }_{\text{similar product recall}}
\end{aligned}
\end{equation}

Using the above reward, we optimize the policy model $\pi_{\theta}$ via GRPO for each target product $b_q$ and its retrieved set $\mathcal{B}_q$. The full objective and notation are provided in Appendix~\ref{sec:grpo_obj}.

\paragraph{Entropy-guided Price Filtering.}
\label{entropy}
Due to sellers' subjective pricing preferences and substantial variation in the condition of retrieved products, the LLM often exhibits considerable uncertainty when generating price suggestions.
To quantify this uncertainty, following recent studies~\cite{geng-etal-2024-survey, fu2025deepthinkconfidence, kang2025scalablebestofnselectionlarge}, we use next-token entropy as a proxy for model confidence, with higher entropy indicating lower confidence.
Formally, at generation step $i$, the entropy $H_i$ of the model's next-token distribution $P_i$ is defined as:
\begin{equation}
    H_i = - \sum_{j \in \mathcal{V}} P_i(j) \log P_i(j),
\end{equation}
where $\mathcal{V}$ is the model's vocabulary, and $P_i(j)$ represents the probability of the $j$-th token in $\mathcal{V}$ at generation step $i$. Since the predicted price $\hat{p}_q$ may span multiple tokens, we estimate its confidence by averaging the entropy across the corresponding generation steps.
Then, an entropy threshold $\theta_\mathcal{H}$ is applied to filter out unreliable price suggestions, retaining only those $\bar{\mathcal{H}} \le \theta_\mathcal{H}$.
In practice, $\theta_\mathcal{H}$ is configured according to application requirements, allowing a flexible trade-off between precision and recall.

\section{Experiments}
\subsection{Settings}
\noindent\textbf{Datasets.}
Our experimental data are collected from real-world e-commerce scenarios. For the test set, we sample 80,000 second-hand products from the top 55 categories, ranked by Gross Merchandise Volume (GMV). After manually filtering out anomalous items, the final test set comprises 72,561 products.
Sourced from the same 55 categories, all products in the training set strictly predate the test samples.
The training data comprises approximately 620,000 samples, with 580,000 allocated to SFT and the remainder for GRPO.
To evaluate generalization, we construct an out-of-distribution (OOD) test set from 611 distinct categories unseen during training (Table~\ref{tab:dataset_statistics_final}).
Standardized products~\cite{li2024commerce} feature uniform brands and model identifiers (e.g., smartphones), whereas non-standardized products are characterized by high variability and a lack of universal specifications.
Further details are provided in Appendix \ref{sec:categorys}.

\begin{table}[htbp]
\centering
\small 

\caption{Statistics of Dataset. ``Cats.'' means Categories.}

\label{tab:dataset_statistics_final}
\setlength{\tabcolsep}{2pt}  
\begin{tabular}{llcc} 
\toprule
\textbf{Split} & \textbf{Subset} & \textbf{TOP 55 Cats.} & \textbf{Other 611 Cats.} \\
\midrule
\multirow{3}{*}{\textbf{Train}} & Standardized & 177,547 & - \\
 & Non-standardized & 438,885 & - \\
 \cmidrule(lr){2-4}
 & \textbf{Total} & \textbf{616,432} & \textbf{-} \\
\midrule
\multirow{3}{*}{\textbf{Test}} & Standardized & 27,947 & 35,670 \\
 & Non-standardized & 44,614 & 174,597 \\
 \cmidrule(lr){2-4}
 & \textbf{Total} & \textbf{72,561} & \textbf{210,267} \\
\bottomrule
\end{tabular}
\end{table}
\noindent\textbf{Metrics.}\label{sec:metrics}
We evaluate accuracy of the price estimation using four metrics: Root Mean Square Log Error (RMSLE) \cite{han2019vision, 9413266}, Mean Absolute Log Error (MALE) \cite{han2019vision, 9413266}, Static Adoption Rate (SAR), and Dynamic Adoption Rate (DAR). 
SAR measures the fraction of predictions whose relative error is within ±20\% of the ground-truth price, whereas DAR employs a dynamic threshold, becoming progressively stricter as the product price increases. 
This property makes it more tolerant of errors for low-value products while imposing tighter accuracy requirements on high-value ones.
\begin{equation}
    \text{RMSLE} = \sqrt{\frac{1}{M} \sum_{i=1}^{M} (\log(\hat{p}_i) - \log(p_i))^2}
\end{equation}
\begin{equation}
    \text{MALE} = \frac{1}{M} \sum_{i=1}^{M} |\log(\hat{p}_i) - \log(p_i)|
\end{equation}
\begin{equation}
    \text{SAR} = \frac{1}{M} \sum_{i=1}^{M} \mathbb{I}\left(\frac{|\hat{p}_i - p_i|}{p_i} \le \tau\right),\tau=0.2
\end{equation}
{\fontsize{10.2pt}{12.24pt}\selectfont
\begin{equation}
    \text{DAR}=\frac{1}{M} \sum_{i=1}^{M} \mathbb{I}\left(\frac{|\hat{p}_i - p_i|}{p_i} \le \tau\right),\;\tau=\frac{a}{\ln(p_i+b)}
\end{equation}
}
Here, $M$ denotes the total number of test samples, while $\hat{p}_i$ and $p_i$ represent the predicted and ground-truth prices for the $i$-th product, respectively. For the DAR metric, we set the hyperparameters $a=1.4$ and $b=10$.

\noindent\textbf{Baselines.}
We compare LLP with the following baselines:\label{xianyuonlinemethod}
\textit{(a) Text-based K-Nearest Neighbors (KNN):} retrieving products using textual representations and estimating the target price as the arithmetic mean of their prices \cite{raykhel2009real};
\textit{(b) Semantic ID (SID)-based KNN:} following the same retrieval-and-averaging procedure using Semantic ID representations;
\textit{(c) Vision-based DNN:} predicting prices using only visual representations \cite{han2019vision};
\textit{(d) Multimodal-based DNN:} leveraging both visual and textual representations \cite{han2020pricesuggestiononlinesecondhand};
\textit{(e) Previous CPV-based Pricing Method:} first identifying the CPV of the target product using TACLR \cite{su2025taclrscalableefficientretrievalbased}, and then estimating its price via a Gaussian Mixture Model over the cluster of products sharing the same CPVs (Figure~\ref{fig:cpv}).
For a fair comparison, we retrieve the top-50 products for all retrieval-based methods.

\noindent\textbf{Models.}
Our experiments employ a diverse selection of LLMs, encompassing both closed-source models like GPT and Claude, as well as leading open-source families, such as the Qwen2.5 \cite{qwen2025qwen25technicalreport} and Qwen3 \cite{yang2025qwen3technicalreport} series. 
We evaluate performance across model scales in Appendix~\ref{sec:scale}.

\noindent\textbf{Parameters.}
We use Qwen2.5-32B-Instruct as the teacher model to construct the training data and Qwen2.5-7B-Instruct as the student model. Our training pipeline performs full-parameter supervised fine-tuning (SFT) on the student model, followed by GRPO, which samples 8 trajectories per prompt. See Appendix~\ref{sec:setting} for more details.

\begin{table*}[htbp]
\centering
\caption{Overall Performance. The best performance is highlighted in \textbf{bold}. "Stan." denotes Standardized Products, while "Non-stan." refers to Non-standardized Products. }
\label{tab:main_result}
\footnotesize 
\setlength{\tabcolsep}{3pt} 
\resizebox{\textwidth}{!}{
\begin{tabular}{l|ccc|ccc|ccc|ccc}
\toprule
\multirow{2}{*}{\textbf{Method}} & \multicolumn{3}{c}{\textbf{RMSLE} $\downarrow$} & \multicolumn{3}{c}{\textbf{MALE} $\downarrow$} & \multicolumn{3}{c}{\textbf{SAR} $\uparrow$} & \multicolumn{3}{c}{\textbf{DAR} $\uparrow$} \\
\cmidrule(lr){2-4} \cmidrule(lr){5-7} \cmidrule(lr){8-10} \cmidrule(lr){11-13}
& Stan. & Non-stan. & Overall & Stan. & Non-stan. & Overall & Stan. & Non-stan. & Overall & Stan. & Non-stan. & Overall \\
\midrule
Text-based KNN & 1.125 & 1.655 & 1.471 & 0.661 & 1.118 & 0.939 & 0.367 & 0.190 & 0.259 & 0.370 & 0.253 & 0.299 \\
SID-based KNN & 0.918 & 1.609 & 1.380 & 0.597 & 1.095 & 0.900 & 0.317 & 0.169 & 0.227 & 0.313 & 0.221 & 0.257 \\
Vision-based DNN & 1.244 & 1.415 & 1.343 & 0.918 & 1.035 & 0.984 & 0.162 & 0.143 & 0.151 & 0.177 & 0.209 & 0.195 \\
Multimodal-based DNN & 0.968 & 1.194 & 1.101 & 0.693 & 0.856 & 0.785 & 0.218 & 0.167 & 0.190 & 0.232 & 0.245 & 0.239 \\
CPV-based Pricing & 0.982 & 1.177 & 1.090 & 0.575 & 0.733 & 0.659 & 0.433 & 0.306 & 0.366 & 0.443 & 0.392 & 0.416 \\
\midrule
Vanilla LLM & 1.244 & 1.589 & 1.463 & 0.932 & 1.227 & 1.111 & 0.197 & 0.158 & 0.174 & 0.198 & 0.192 & 0.195 \\
LLM (w/ SFT) & 1.749 & 1.288 & 1.486 & 1.341 & 0.916 & 1.082 & 0.173 & 0.242 & 0.215 & 0.184 & 0.298 & 0.253 \\
LLM (w/ Retrieval) & 0.487 & 0.841 & 0.723 & 0.272 & 0.510 & 0.417 & 0.610 & 0.417 & 0.492 & 0.624 & 0.513 & 0.557 \\
LLM (w/ Ret.+SFT) & \textbf{0.442} & 0.744 & 0.643 & 0.244 & 0.459 & 0.375 & 0.649 & 0.459 & 0.534 & 0.664 & 0.550 & 0.594 \\
\midrule
\rowcolor{gray!20}
LLP (Ours) & 0.459 & \textbf{0.714} & \textbf{0.627} & \textbf{0.239} & \textbf{0.429} & \textbf{0.355} & \textbf{0.652} & \textbf{0.478} & \textbf{0.546} & \textbf{0.669} & \textbf{0.572} & \textbf{0.610} \\
\bottomrule
\end{tabular}
}
\end{table*}
\subsection{Results}
As demonstrated in Table~\ref{tab:main_result}, our proposed LLP significantly outperforms all baselines on the second-hand price prediction task, highlighting its superior performance.
Specifically, 
benefiting from the robust semantic comprehension of LLMs, LLP significantly outperforms the simplistic mean-based approach, KNN, given the same retrieved items.
Both the vanilla LLM and its SFT variant exhibit poor performance due to a lack of domain knowledge and overfitting, respectively.
In contrast, incorporating retrieved products as supplementary context leads to marked performance gains.
For instance, the SAR on standardized products increases
more than threefold (from 19.7\% to 61.0\%), underscoring the importance of incorporating market references. Building on retrieval augmentation, SFT and GRPO further improve pricing performance, yielding a nearly 20\% gain in DAR over the strongest CPV-based baseline. These results validate the effectiveness of our LLP.

\subsection{Analysis}
\subsubsection{Generalization Performance}
\begin{table}[htbp]
\centering
\footnotesize

\caption{Analysis on 611 OOD Categories.}

\label{tab:generalization_performance}
\setlength{\tabcolsep}{2pt}
\resizebox{\linewidth}{!}{
\begin{tabular}{lcccc} 
\toprule
\textbf{Method} & \textbf{RMSLE} & \textbf{MALE} & \textbf{SAR} & \textbf{DAR} \\
\midrule
Multimodal-based DNN  & 1.439 & 1.103 & 0.123 & 0.181 \\\midrule
LLM (w/ SFT)  & 5.014 & 5.014 & 0.060 & 0.061 \\
LLM (w/ Retrieval) & 0.938 & 0.581 & 0.346 & 0.438 \\
LLM (w/ Retrieval+SFT) & 0.854 & 0.539 & 0.394 & 0.484 \\\midrule
\rowcolor{gray!20}
LLP (w/ Retrieval+SFT+GRPO) & \textbf{0.808} & \textbf{0.487} & \textbf{0.429} & \textbf{0.523} \\
\bottomrule
\end{tabular}
}
\end{table}
To assess generalizability, we evaluate LLP on another 611 categories. As shown in Table~\ref{tab:generalization_performance}, LLP significantly outperforms all baselines.
Both the multimodal-based DNN and fine-tuned LLMs fail to generalize to unseen categories, with the latter achieving a SAR of merely 6\%. In contrast, integrating product retrieval yields a substantial boost, elevating the SAR to 34.6\%.
Building on this improvement, SFT raises SAR to 39.4\%, and GRPO further improves it to 42.9\%. These consistent gains demonstrate that LLP generalizes well to categories unseen during training.

\subsubsection{Entropy-guided Filtering Analysis}
\begin{figure}[htbp]
    \centering
    \includegraphics[width=\columnwidth]{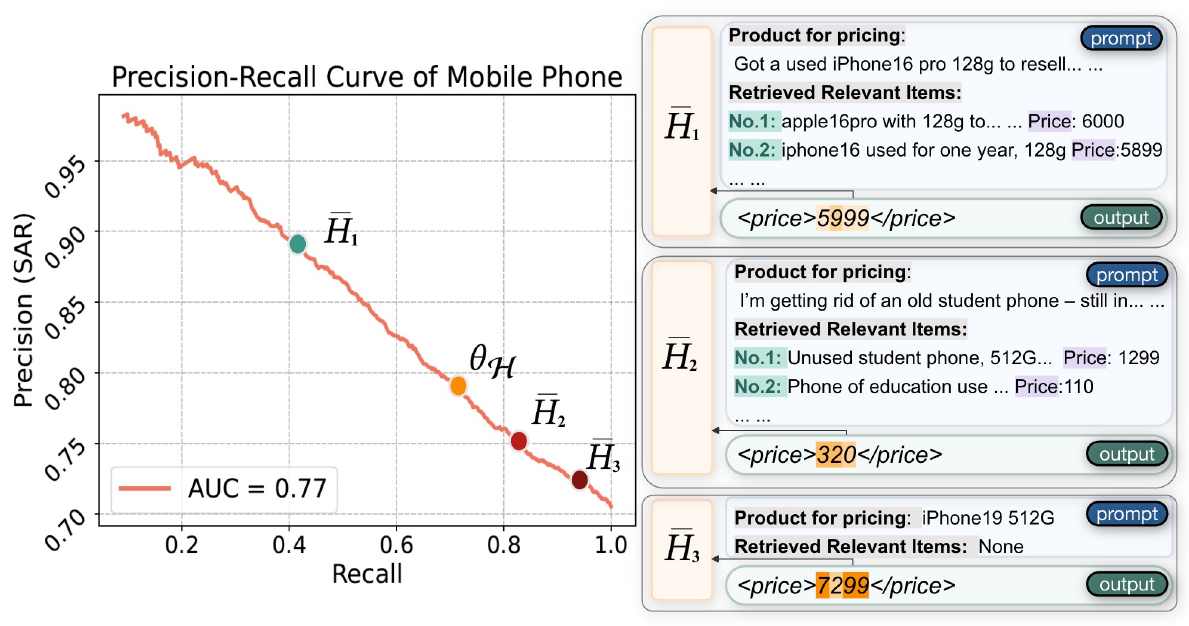}
    \caption{Entropy-based PR Curve of the Mobile Phone Category. Darker orange shading signifies higher token entropy. $\bar H_1$: well-described product (low entropy); $\bar H_2$: ambiguous input lacking key attributes; $\bar H_3$: unreleased product with no retrieved references (high entropy).}
    \label{fig:entropy_image}
\end{figure}
To validate the effectiveness of our entropy-based price filtering, we conducted a case study on the mobile phone category. Figure~\ref{fig:entropy_image} illustrates the Precision-Recall (PR) curve obtained by varying the entropy threshold.
At the threshold $\theta_\mathcal{H}$, LLP achieves a high precision of 78\% at 70\% product coverage.
Ambiguous inputs (e.g., $\bar{H_2}$) that lack key attributes (e.g., brand and model), result in higher output entropy, thereby automatically triggering the price filtering.
For unreleased products, the failure to retrieve similar products leads to maximal pricing uncertainty (e.g., $\bar{H}_3$).
With an AUC of 0.77, the entropy score provides a reliable confidence signal, enabling a flexible precision-recall trade-off to meet different requirements in real-world deployment. Further analyses are provided in Appendix~\ref{sec:entropy}.

\subsection{Case Study}
\begin{table}[htbp]
\centering
\small
\setlength{\tabcolsep}{3pt}
\caption{Case study on \textit{Apple iPhone 16, 256\,GB} under different battery conditions.}
\label{tab:case_study}
\begin{tabular}{@{}lccc@{}}
\toprule
\textbf{Product} & \textbf{Battery Condition} & \textbf{CPV (\$)} & \textbf{LLP (\$)} \\
\midrule
iPhone 16, 256GB & Battery Health 95\% & 870 & 840 \\
iPhone 16, 256GB & Battery Health 70\% & 870 & 780 \\
iPhone 16, 256GB & Battery dead        & 870 & 653 \\
\bottomrule
\end{tabular}
\end{table}
\label{sec:case_study}
To provide an intuitive illustration of LLP's advantage over the CPV-based baseline, we present a case study of three second-hand listings of the same brand, model and storage (\textit{Apple iPhone 16, 256\,GB}) that differ only in their battery condition, as shown in Table~\ref{tab:case_study}. The CPV-based method maps the three products to a single cluster and assigns them a uniform price of 870, ignoring fine-grained condition differences. 
In contrast, LLP captures variations in battery health and predicts progressively lower prices as it deteriorates. For a device with a dead battery, LLP predicts 653, which is 217 lower than the CPV-based method’s unchanged estimate of 870.
This case demonstrates that LLP provides more condition-aware and appropriate price suggestions than the CPV-based baseline.

\subsection{Online Deployment}
To evaluate the performance of our pricing system, we conduct a 10-day online A/B test, with the CPV-based method and LLP each receiving 20\% of platform traffic. Compared with the CPV-based method, LLP raises \textit{direct adoption}---the proportion of sellers who submit the suggested price without modification---from 6\% to 17\%. Consistent with this improvement, online SAR, which measures the proportion of suggestions falling within $\pm20\%$ of the seller's final submitted price, also increases from 40\% to 72\%. By helping sellers set more reasonable prices, LLP delivers a 0.7\% lift in GMV and a 1.4\% gain in Click-Through Conversion Rate (CTCVR) during the A/B test. LLP has been fully deployed in production on 20 NVIDIA L20 GPUs, supporting 25 QPS with an end-to-end latency of approximately 790\,ms. The detailed latency breakdown is provided in Appendix~\ref{sec:latency}.

\section{Related Works}
C2C e-commerce has flourished in recent years. However, setting appropriate prices for second-hand products remains challenging for inexperienced individual sellers.
Numerous studies have explored intelligent pricing methods to improve listing efficiency, a key objective of our work.
Early research focuses on traditional machine learning algorithms \cite{raykhel2009real, 8022758, 9740817}. However, these methods are tailored to specific products, limiting their generalization across diverse categories.
Subsequent research shifts to deep neural networks, modeling complex item-price mappings.
While \citet{7937942} and \citet{han2019vision} rely on single-modal features, \citet{han2020pricesuggestiononlinesecondhand} integrate both visual and textual information to improve the accuracy of price estimation. 
More recently, \citet{vedula2025quantileregressionlargelanguage} incorporate quantile regression into LLMs to predict price distributions. In contrast, LLP recasts price prediction as a generative task that captures market dynamics and analyzes fine-grained differences in product condition, effectively addressing key limitations of prior works. 

\section{Conclusion}
In this paper, we present LLP, an LLM-based generative pricing system for C2C e-commerce platforms. Inspired by the Law of One Price, LLP retrieves similar products as dynamic market references and leverages LLMs' fine-grained semantic understanding of colloquial product descriptions, effectively overcoming the poor generalization of prior regression-based models. During training, it enhances reasoning faithfulness and pricing accuracy through bidirectional reasoning-based supervision and retrieval-aware policy optimization. During inference, it uses entropy as a confidence signal to filter out unreliable price suggestions. Extensive offline and online evaluations demonstrate the effectiveness of our method.

\section*{Limitations}
LLP establishes a novel generative paradigm for intelligent pricing of second-hand products on C2C e-commerce platforms. Although LLP employs multimodal Semantic IDs for product representation during retrieval, its reasoning stage currently relies solely on textual descriptions. Product images provide rich visual cues for identifying key attributes such as category, brand, model, and condition. Consequently, extending LLP with multimodal large language models (MLLMs) to leverage this visual information is a promising direction. We acknowledge this limitation. Given the substantial inference cost and latency of MLLMs, future work will explore image-token pruning to enable efficient multimodal reasoning and deployment.

\bibliography{custom}

\clearpage
\appendix
\section{Appendix}
\label{sec:appendix}

\subsection{Online Serving}
\label{sec:latency}
We further analyze the online latency of LLP in comparison with the previous CPV-based pricing method. Table~\ref{tab:latency} reports a detailed latency breakdown, including the retrieval stage and the inference stage of LLMs. The CPV-based method first predicts the Category Property Values (CPVs) of the target product to map it to a pre-computed cluster, and then estimates the price via a Gaussian Mixture Model (GMM), with an end-to-end latency of about 450 ms. Although LLP introduces additional computation compared with the CPV-based method, its average end-to-end latency remains below 800\,ms, satisfying the latency requirements of our online service.

\begin{table}[htbp]
\centering
\small
\caption{Latency comparison (ms) between LLP and the CPV-based pricing method.}
\label{tab:latency}
\begin{tabular}{lccc}
\toprule
\textbf{Method} & \textbf{Retrieval} & \textbf{Inference} & \textbf{Total} \\
\midrule
CPV-based Pricing & -- & -- & 450 \\
LLP               & 40 & 750 & 790 \\
\bottomrule
\end{tabular}
\end{table}

\subsection{Prompt Sensitivity}
\label{sec:prompt_sensitivity}
To evaluate the robustness of LLP to prompt design, we conduct a sensitivity analysis across four prompt variants. Specifically, we consider: (A) the original prompt used in our paper; (B) a rephrased version that substitutes instructions with synonymous expressions; (C) an ordering variant that swaps the chain-of-thought and the predicted price in the output; and (D) a variant that randomly shuffles the order of the retrieved products. The results are reported in Table~\ref{tab:prompt_sensitivity}.

Variant~B performs on par with the original, suggesting that LLP is insensitive to paraphrasing of the instructions. Variant~C yields a larger drop, as chain-of-thought can improve reasoning by generating intermediate steps prior to the final answer \cite{wei2022chain}.
Variant~D, shuffling the retrieved products, incurs the largest drop. The retrieval ranking itself encodes similarity to the target product, but random shuffling discards this signal.
Despite these effects, the overall SAR fluctuation remains within 1.2\%, demonstrating that LLP is robust to the prompt formulation.
\begin{table}[htbp]
\centering
\small
\setlength{\tabcolsep}{5pt}
\caption{Prompt sensitivity analysis of LLP. $\Delta$ denotes the absolute change in SAR relative to Variant~A.}
\label{tab:prompt_sensitivity}
\begin{tabular}{@{}clcc@{}}
\toprule
\textbf{Variant} & \textbf{Description} & \textbf{SAR} & \textbf{$\Delta$} \\
\midrule
A & Original (paper)            & \textbf{0.546} & --      \\
B & Rephrased instructions      & 0.545          & $-0.001$ \\
C & Swapped reasoning and price     & 0.538          & $-0.008$ \\
D & Shuffled retrieved products & 0.535          & $-0.011$ \\
\bottomrule
\end{tabular}
\end{table}

\subsection{Statistical Significance}
\label{sec:significance}
To assess LLP's SAR improvement over the CPV-based baseline, we conduct a significance test on 72,561 samples. As shown in Table~\ref{tab:significance}, the 95\% CIs are narrow (approximately one percentage point wide) and non-overlapping. Together with a $p$-value below $0.001$, these results indicate that LLP's SAR gain is statistically significant and unlikely to be caused by random sampling variation.

\begin{table}[htbp]
\centering
\small
\caption{Statistical significance analysis of LLP against the CPV-based baseline on 72{,}561 test samples.}
\label{tab:significance}
\begin{tabular}{lccc}
\toprule
\textbf{Method} & \textbf{SAR} & \textbf{95\% CI} & \textbf{$p$-value} \\
\midrule
CPV-based   & 0.3656 & [0.3607, 0.3705] & -- \\
LLP (Ours)  & 0.5462 & [0.5414, 0.5510] & $p < 0.001$ \\
\bottomrule
\end{tabular}
\end{table}

\subsection{Error Analysis}
\label{sec:error_analysis}
To better understand the limitations of LLP in real-world deployment, we collect online failure cases and manually categorize each case into one of four types, as summarized in Table~\ref{tab:error_analysis}: (1) \textit{\textbf{Pricing Subjectivity}} accounts for roughly 40\% of the failure cases, stemming from the inherent subjectivity of individual sellers (e.g., urgent sellers tend to price lower, whereas patient sellers tend to price higher than the market reference); (2) \textit{\textbf{Incomplete Information}} contributes about 20\%, where user-submitted product descriptions lack key attributes such as brand, model, or condition, making accurate pricing infeasible; (3) \textit{\textbf{Retrieval Quality}} is responsible for another 20\%, caused by either failing to retrieve relevant products or retrieving similar products with excessively large price variance; (4) \textit{\textbf{Reasoning Errors}} account for the remaining 20\%, where LLP incorrectly analyzes the retrieved products or generates hallucinated conclusions.
These observations suggest that future improvements could focus on richer user-side information elicitation, more robust retrieval, and more faithful LLM-based reasoning.

\begin{table}[htbp]
\centering
\small
\setlength{\tabcolsep}{4pt}
\renewcommand{\arraystretch}{1.15}
\caption{Error analysis of LLP on online failure cases.}
\label{tab:error_analysis}
\begin{tabular}{@{}>{\raggedright\arraybackslash\hyphenpenalty=10000\exhyphenpenalty=10000}p{0.30\linewidth}c>{\raggedright\arraybackslash}p{0.50\linewidth}@{}}
\toprule
\textbf{Error Type} & \textbf{Prop.} & \textbf{Description} \\
\midrule
Pricing Subjectivity   & \textbf{40\%} & Inherent seller subjectivity, leading to divergent price expectations for similar items. \\
Incomplete Information & \textbf{20\%} & User-submitted description lacking key attributes, such as brand, model, and condition. \\
Retrieval Quality      & \textbf{20\%} & No relevant product retrieved, or retrieved products exhibit large price variance. \\
Reasoning Errors       & \textbf{20\%} & LLP incorrectly analyzes the retrieved products or generates hallucinations. \\
\bottomrule
\end{tabular}
\end{table}

\subsection{Human Evaluation}
\label{sec:human_eval}
\begin{table*}[htbp]
\centering
\footnotesize
\caption{Human evaluation of 1,000 forward-reasoning trajectories randomly sampled from the training data. The first three criteria assess rationale quality, while the remaining three assess rationale reliability.}
\label{tab:human_evaluation}
\setlength{\tabcolsep}{3pt}
\renewcommand{\arraystretch}{1.08}
\begin{tabular}{@{}>{\raggedright\arraybackslash}p{0.32\textwidth}>{\raggedright\arraybackslash}p{0.58\textwidth}c@{}}
\toprule
\textbf{Metric} & \textbf{Criterion} & \textbf{Result} \\
\midrule
Evidence Fidelity~$\uparrow$ & The rationale is accurate and supported by the product information and retrieved evidence. & 96\% \\
Comparable Relevance~$\uparrow$ & The referenced products are appropriate comparisons for the target product on key attributes. & 94\% \\
Price Justification~$\uparrow$ & The suggested price is reasonably justified by the retrieved products. & 99\% \\
\addlinespace[2pt]
No Explicit Target-price Leakage~$\uparrow$ & The rationale does not explicitly reveal or refer to the known price of the target product. & 92\% \\
No Backward Reasoning Disclosure~$\uparrow$ & The rationale does not mention the backward-reasoning process or the selected subset. & 96\% \\
No Observable Post-hoc Rationalization~$\uparrow$ & The rationale does not justify an exact price without sufficient support from the retrieved products. & 98\% \\
\bottomrule
\end{tabular}
\end{table*}

To assess the quality and reliability of rationales synthesized during the forward-reasoning stage, we manually audit 1,000 randomly sampled trajectories. As shown in Table~\ref{tab:human_evaluation}, all six evaluation criteria achieve high pass rates. In particular, 98\% of the trajectories show no observable post-hoc rationalization, providing evidence for the faithfulness of the synthesized rationale supervision. The results also indicate that the vast majority of forward-reasoning trajectories do not leak privileged information, such as the ground-truth price. Accounting for overlap across issue categories, only 9\% of the trajectories are identified as problematic. The low rate supports the use of synthesized rationales as reliable supervision for student post-training. During real-world inference, the student has access to neither the ground-truth price nor any other privileged information. Moreover, GRPO goes beyond direct teacher imitation by sampling multiple trajectories per prompt and reinforcing the best-performing ones, thereby further mitigating the issues described above.

\subsection{Discussion on Pricing Bias}
\label{sec:bias}
We further discuss potential biases in LLP from two perspectives: the user side and the category side.
 (1) \textbf{\textit{From the user perspective}}, LLP generates price suggestions solely based on product-side attributes (e.g., brand, model, condition, and specifications), without relying on any user-related information. Moreover, all suggested prices are advisory only. Individual sellers may freely adopt, modify, or reject the suggested price.
 (2) \textbf{\textit{From the category perspective}}, we find that overestimations and underestimations are roughly symmetric across all categories, with an imbalance within 5\% per category, indicating that LLP does not exhibit systematic category-level pricing bias. Together, these findings confirm that it delivers user-agnostic and category-balanced price suggestions.

\subsection{GRPO Objective}
\label{sec:grpo_obj}
For each target product $b_q$ and its retrieved context $\mathcal{B}_q$, GRPO samples a group of outputs $\{o_1, o_2, \cdots, o_G\}$ from the old policy $\pi_{\theta_{old}}$. Based on the rewards defined in Section~\ref{grpo}, the policy model $\pi_{\theta}$ is optimized by maximizing the following objective:
\begin{equation}
\small
\begin{split}
    &J_{\text{GRPO}}(\theta) ={} \mathbb{E}_{x \sim \mathcal{D},\, \{o_i\}_{i=1}^G \sim \pi_{\theta_{\text{old}}}(\cdot|x)} \bigg[ \frac{1}{G} \sum_{i=1}^G \Big(\min\big(\\ & w_i(\theta)A_i,
    \hphantom{\mathbb{E}_{}} \text{clip}(w_i(\theta), 1-\epsilon, 1+\epsilon)A_i \big) - \beta\, \mathbb{D}_{KL} \Big) \bigg]
\end{split}
\end{equation}
Here, $w_i(\theta)=\frac{\pi_{\theta}(o_i|x)}{\pi_{\theta_{old}}(o_i|x)}$ is the importance weight, with $\epsilon$ and $\beta$ as hyperparameters. $A_i$ is the advantage computed by group-wise normalization of the rewards $\{r_1, r_2, \cdots, r_G\}$:
\begin{equation}
\small
    A_i = \frac{r_i - \mathrm{mean}(\{r_j\}_{j=1}^{G})}{\mathrm{std}(\{r_j\}_{j=1}^{G})}
\end{equation}
$\mathbb{D}_{KL}$ is the KL divergence that penalizes excessive deviations of the current policy from the reference model $\pi_{ref}$.

\subsection{Experimental Setup}
\label{sec:setup}

\paragraph{Dataset}
\label{sec:categorys}
In e-commerce, products can be broadly classified into two categories based on their degree of standardization: standardized and non-standardized products. Standardized products refer to items with clear specifications, models, and functions that typically adhere to recognized industry standards. Non-standardized products, conversely, are defined by an emphasis on individual attributes and subjective appeal rather than uniform standards. Consequently, consumer decisions are driven more by factors such as style, design, and brand identity, leading to lower price sensitivity.
Specific categories for both standardized and non-standardized items are listed in Table \ref{tab:category_examples}.

\begin{table*}[t]
\centering
\small
\sloppy
\caption{Examples of Standardized and Non-standardized Categories.}
\label{tab:category_examples}
\setlength{\tabcolsep}{4pt}
\begin{tabular}{@{}lp{0.82\linewidth}@{}}
\toprule
\textbf{Classification} & \textbf{Category} \\
\midrule
\textbf{Standardized} & Graphics Cards, Motorcycles, Game Consoles, Automobiles, Mobile Phones, Tablet PCs, Billiard Cues, Digital Cameras, Laptops, Smartwatches, Electric Bicycles, Custom-built Desktops, Air Conditioners, Bluetooth Earphones, Fishing Rods, Keyboards \\
\midrule
\textbf{Non-standardized} & Dresses, Handbags, T-shirts, Watches, Action Figures (GKs), Coffee/Tea, Low-top Shoes, Necklaces, Running Shoes, Dolls, Bracelets, Lolita Fashion, Skate Shoes, Cotton Dolls, Concert Tickets, Video Memberships, Phone Components, Gimbals, E-bike Batteries, Motorcycle Brake Systems, E-bike Controllers, Hotel Booking Services, Attraction Tickets, Buffet Vouchers, Proxy Shopping Services, Food Coupons, Fast Food Vouchers, Mobile Top-ups \\
\bottomrule
\end{tabular}
\end{table*}

\begin{figure*}[htbp]
  \centering
  \includegraphics[width=\textwidth]{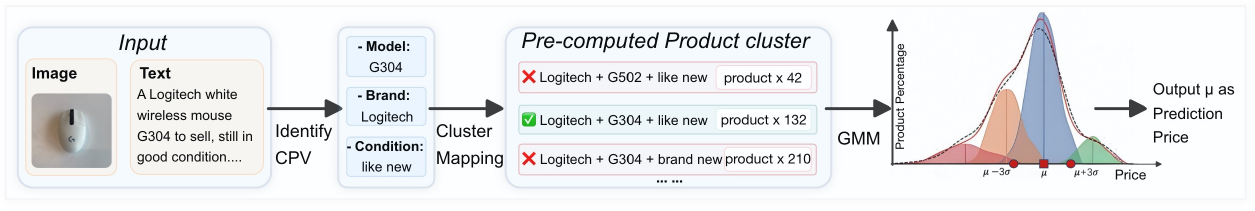}
  \caption{Previous CPV-based Pricing Framework for C2C E-commerce Platforms.}
  \label{fig:cpv}
\end{figure*}

\paragraph{Implementation Details}
\label{sec:setting}
The detailed experimental settings are as follows:
\textbf{\textit{1) Bidirectional Reasoning-based Fine-Tuning.}}
We first synthesize the training data using Qwen2.5-32B-Instruct served by vLLM, with greedy decoding (temperature $= 0$) and a maximum context length of 8{,}192 tokens.
We then fine-tune Qwen2.5-7B-Instruct for 4{,}000 steps with a learning rate of $1 \times 10^{-5}$, a per-device batch size of 2, and 4 gradient accumulation steps. The training set is a 1:1 mixture of samples with and without the reasoning process $R$, distinguished via system prompts.
\textbf{\textit{2) Retrieval-aware Policy Optimization. }} Following supervised fine-tuning, we employ Group Relative Policy Optimization (GRPO) to further optimize the model's output distribution. To facilitate exploration and ensure diversity within the sampled groups, we set the sampling temperature to $0.8$ and top-p to $0.95$. We train for 1,000 steps, sampling $G=8$ outputs per prompt, with a maximum generation length of 512 tokens. The optimization is constrained by a KL coefficient of 0.04 and a clip range of $\epsilon=0.2$. In the outcome reward, the sensitivity coefficient is set to $\alpha=25$, which assigns a reward of $0.5$ at a relative price error of $20\%$.
For evaluation, we use greedy decoding to guarantee reproducibility.
All experiments are conducted on 64 NVIDIA H20 GPUs. 

\subsection{Additional Analysis}

\paragraph{Analysis on Various LLMs}
\label{sec:scale}

Table~\ref{tab:foundation_model_comparison} provides a detailed performance comparison of various open-source and closed-source LLMs on the task of price prediction for second-hand products. All models reported in Table~\ref{tab:foundation_model_comparison} are \textit{off-the-shelf} LLMs without any fine-tuning, and use exactly the same retrieved products as in Table~\ref{tab:main_result}. The goal of this experiment is to examine how pricing performance varies across LLMs of different scales, thereby informing the choice of a base model that balances accuracy against inference cost in deployment. Since this serves as a supplementary analysis rather than a core comparison, we present it here.

We observe a clear Scaling Law: as the number of model parameters increases, the accuracy of the LLMs' price prediction also improves. The performance improvement slows down when the model size grows from 14B to 72B, with diminishing marginal returns. For models of comparable scale, the Qwen2.5 series consistently outperform Qwen3 series, for which the thinking mode is disabled to mitigate inference latency. Furthermore, the experimental results indicate that the open-source Qwen2.5-72B-Instruct outperforms GPT-4.1-mini and Claude 3.5-Haiku across all metrics, and remains competitive with the latest proprietary models such as GPT-5.2 and Gemini-2.5-Pro. Consequently, we ultimately select the Qwen2.5 series for pricing tasks.

We note that scaling up the base model can yield additional accuracy gains, but the associated inference overhead grows substantially. In real-world deployment, our system serves millions of individual sellers per day, imposing stringent constraints on serving cost and latency. Balancing these practical considerations with pricing accuracy, we adopt Qwen2.5-7B-Instruct as the base model of LLP. Notably, after our proposed post-training, the resulting 7B model achieves lower pricing error than the vanilla Qwen2.5-72B-Instruct and even closed-source counterparts such as GPT-5.2 and Gemini-2.5-Pro, while being orders of magnitude cheaper to serve.

\begin{table}[htbp]
\centering
\small
\caption{Performance Comparison of Various Models. The best performance is highlighted in \textbf{bold}.}
\label{tab:foundation_model_comparison}
\setlength{\tabcolsep}{1pt}
\setlength{\fboxsep}{2pt}
\newcommand{\hlbest}[1]{\colorbox{gray!20}{#1}}
\begin{tabular}{lcccc}
\toprule
\textbf{Model} & \textbf{RMSLE} $\downarrow$ & \textbf{MALE} $\downarrow$ & \textbf{SAR} $\uparrow$ & \textbf{DAR} $\uparrow$ \\
\midrule
\multicolumn{5}{c}{\textit{Open-source Models}} \\
\midrule
Qwen2.5-3B-Instruct & 0.881 & 0.503 & 0.434 & 0.489 \\
Qwen2.5-7B-Instruct & 0.723 & 0.417 & 0.492 & 0.557 \\
Qwen2.5-14B-Instruct & 0.659 & 0.381 & 0.511 & 0.573 \\
Qwen2.5-32B-Instruct & 0.666 & 0.382 & 0.503 & 0.565 \\
Qwen2.5-72B-Instruct & \hlbest{\textbf{0.627}} & 0.364 & 0.523 & 0.586 \\
Qwen3-4B & 0.805 & 0.470 & 0.417 & 0.471 \\
Qwen3-8B & 0.718 & 0.412 & 0.483 & 0.542 \\
Qwen3-14B & 0.667 & 0.386 & 0.487 & 0.546 \\
Qwen3-32B & 0.639 & 0.387 & 0.476 & 0.533 \\
Qwen3-30B-A3B & 0.739 & 0.405 & 0.490 & 0.547 \\
\midrule
\multicolumn{5}{c}{\textit{Closed-source Models}} \\
\midrule
Claude 3.5-Haiku & 0.812 & 0.449 & 0.469 & 0.524 \\
GPT-4.1-mini & 0.676 & 0.378 & 0.521 & 0.582 \\
GPT-5.2 & 0.662 & 0.375 & \hlbest{\textbf{0.534}} & \hlbest{\textbf{0.599}} \\
Gemini-2.5-Pro & 0.649 & \hlbest{\textbf{0.360}} & 0.526 & 0.598 \\
\bottomrule
\end{tabular}
\end{table}

\paragraph{Analysis on Different Product Representation}
\begin{table}[htbp]
\centering
\small 
\caption{Performance Comparison of Different Product Representations During the Retrieval Phase.}
\label{tab:modality_ablation}
\setlength{\tabcolsep}{2pt}
\begin{tabular}{lcccc}
\toprule
\textbf{Method} & \textbf{RMSLE} $\downarrow$ & \textbf{MALE} $\downarrow$ & \textbf{SAR} $\uparrow$ & \textbf{DAR} $\uparrow$ \\
\midrule
Text  & 0.9982 & 0.5557 & 0.4166 & 0.4719 \\
Image  & 0.8817 & 0.5085 & 0.4276 & 0.4802 \\
\rowcolor{gray!20}
Semantic ID (Ours) & \textbf{0.7234} & \textbf{0.4167} & \textbf{0.4923} & \textbf{0.5569} \\
\bottomrule
\end{tabular}
\end{table}
In this section, we investigate the impact of different product representations on the LLMs' price prediction. We conduct this analysis using the vanilla Qwen2.5-7B-Instruct for all experiments. Table \ref{tab:modality_ablation} shows that retrieval based on product image representations leads to better performance compared to text representations. This suggests that product images carry essential information as well. More importantly, the multimodal Generative Semantic ID representation consistently outperforms both unimodal alternatives across all metrics, reducing RMSLE by 18.0\% and 27.5\% compared with image- and text-based retrieval, respectively.

\paragraph{Analysis on the Number of Retrieved Products}
\begin{table}[htbp]
\centering
\small
\caption{Performance Comparison with a Varying Number of Retrieved Products (k). The best performance is highlighted in \textbf{bold}.}
\label{tab:rag_sensitivity}
\setlength{\tabcolsep}{4pt}
\begin{tabular}{ccccc}
\toprule
\textbf{Number (k)} & \textbf{MALE} $\downarrow$ & \textbf{RMSLE} $\downarrow$ & \textbf{SAR} $\uparrow$ & \textbf{DAR} $\uparrow$ \\
\midrule
0  & 1.0912 & 1.5156 & 0.1923 & 0.2368 \\
1  & 0.4872 & 0.8792 & 0.4569 & 0.5160 \\
3  & 0.4175 & 0.7277 & 0.4945 & 0.5568 \\
5  & 0.3972 & 0.6952 & 0.5086 & 0.5725 \\
15 & 0.3702 & 0.6502 & 0.5327 & 0.5964 \\
30 & 0.3598 & 0.6348 & 0.5419 & 0.6068 \\
\rowcolor{gray!20}
50 & \textbf{0.3549} & \textbf{0.6263} & \textbf{0.5462} & \textbf{0.6110} \\
100 & 0.3584 & 0.6328 & 0.5414 & 0.6099 \\
\bottomrule
\end{tabular}
\end{table}
To study the effect of the number of retrieved products on price prediction, we perform an experiment with the results shown in Table \ref{tab:rag_sensitivity}. Overall, as the number of retrieved products varies from 0 to 100, the performance of our LLP first increases and then declines. Nevertheless, leveraging retrieved similar products as a basis for reasoning consistently outperforms LLMs without external knowledge, which validates the effectiveness of the retrieval module. Initially, with a small number of retrieved products, the LLMs' price prediction accuracy is constrained by a lack of relevant information, so it rises as more products are added. The performance drops when retrieving a large quantity of products (e.g., 100). This can be attributed to interference from irrelevant products, and the LLMs' sensitivity to token positions \cite{liu-etal-2024-lost, xu2024retrieval, jin2025longcontext, peysakhovich2023attention, qin-etal-2023-nlp}, a phenomenon known as position bias, which arises from the position encoding.
Accordingly, we select 50 products as the market references in our practical deployment, which yields the best pricing performance while maintaining acceptable inference overhead.

\begin{figure}[htbp]
    \centering
    \begin{subfigure}{0.492\columnwidth}
        \centering
        \includegraphics[width=\linewidth]{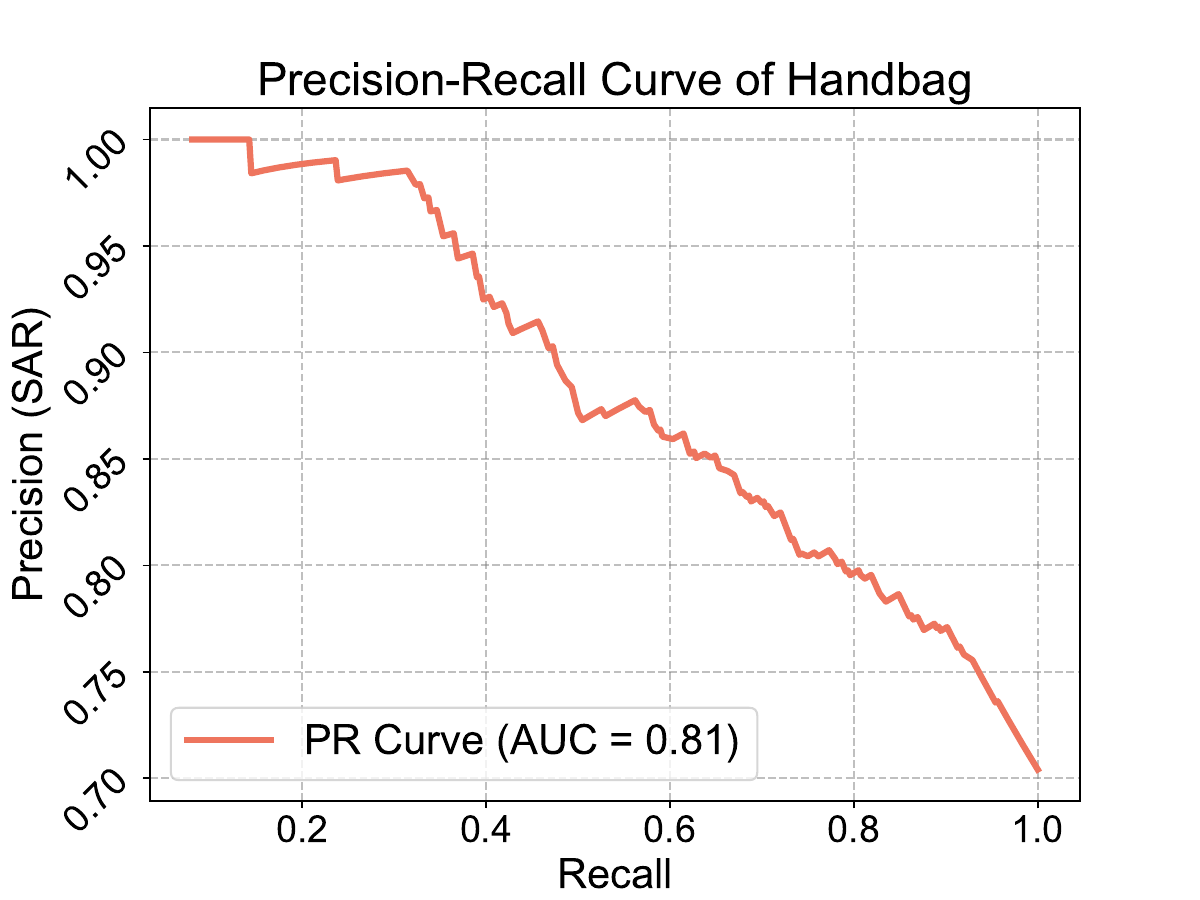}
        \caption{Handbag Category}
        \label{fig:handbag}
    \end{subfigure}
    \begin{subfigure}{0.492\columnwidth}
        \centering
        \includegraphics[width=\linewidth]{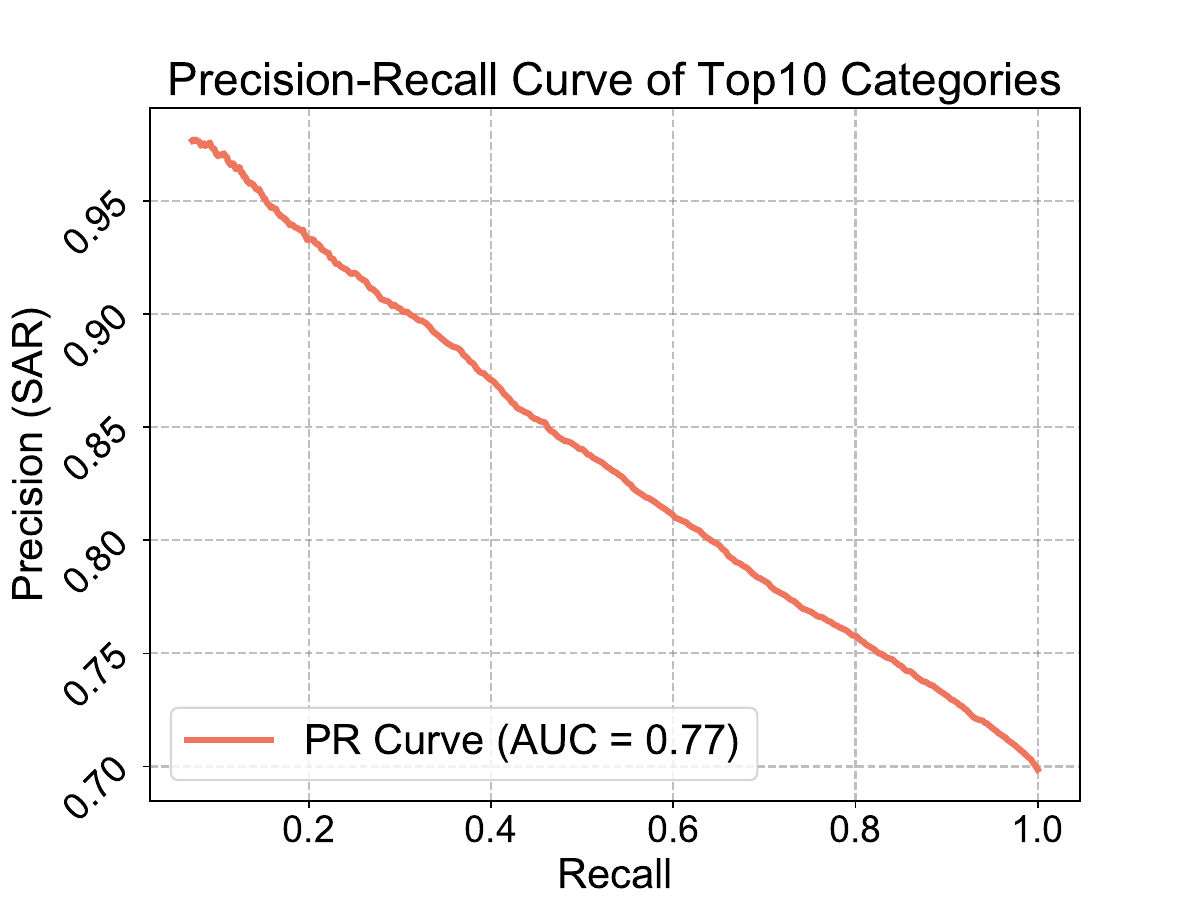}
        \caption{Top 10 Category}
        \label{fig:top10}
    \end{subfigure}
    \caption{Entropy-based PR Curve of the Handbag and Top 10 Categories.}
    \label{fig:pr_categroy}
\end{figure}

\paragraph{Analysis on Entropy-based Price Filtering}
\label{sec:entropy}
Figure \ref{fig:pr_categroy} further shows the Precision-Recall (PR) curve for handbags and the top 10 categories, obtained by dynamically tuning the entropy threshold. The trends are similar to those observed in the mobile phone category, demonstrating the approach's generality and robustness across diverse product categories. On the top 10 categories, it achieves over 80\% pricing accuracy at 60\% product coverage, demonstrating strong reliability in real-world deployment. Even under full coverage, where every product can be priced, LLP still sustains a pricing accuracy near 70\%. In the single handbag category, at 20\% product coverage, SAR approaches 100\%, demonstrating high pricing accuracy. At the expense of partial product coverage, the system delivers a dramatically improved user experience, while enabling a dynamic trade-off between precision and recall.

\subsection{Semantic ID}
\label{sec:semantic}

\begin{figure}[htbp]
    \centering
    \includegraphics[width=\columnwidth]{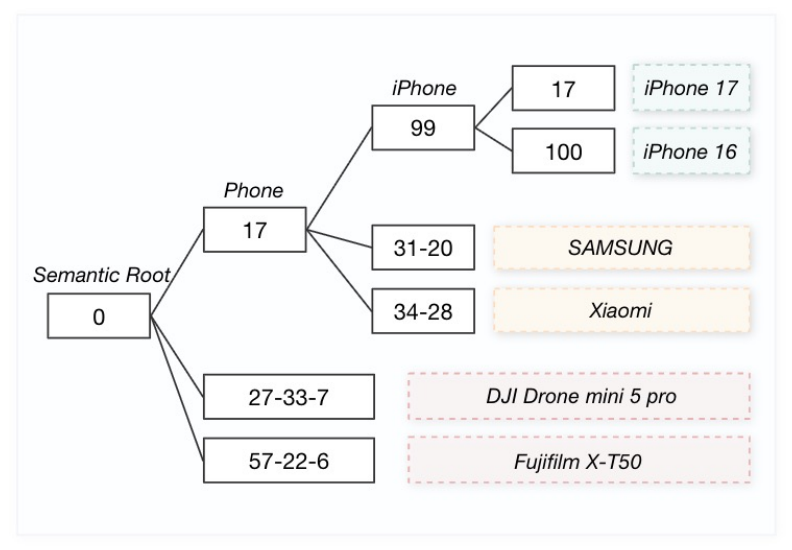}
    \caption{Examples of Semantic IDs.}
    \label{fig:gsid1}
\end{figure}

Following the generative tokenization learning paradigm, we adopt a T5-based model that autoregressively transforms products into hierarchical Semantic IDs via coarse-to-fine progressive quantization. Built upon step-specific codebooks, we construct a generative tree structure that integrates multimodal features and collaborative signals, facilitating knowledge transfer and ensuring robust generalization (as illustrated in Figure~\ref{fig:gsid1}).

\paragraph{Encoder--Decoder Architecture.}
We maintain $M$ step-specific codebooks $\{E_t\}_{t=1}^{M}$, each $E_t\!\in\!\mathbb{R}^{K\times D}$, where each row of $E_t$ stores the embedding of a discrete code at level $t$. Given a product description $b$, the encoder extracts its semantic representation, while the decoder autoregressively generates Semantic IDs conditioned on the encoder output and the embeddings of previously produced codes. At step $t$, the decoder yields a hidden state
\begin{equation}
\small
    d_t = \mathrm{Decoder}\!\Big(\mathrm{Encoder}(b),\, \big(E_\tau[z_\tau]\big)_{\tau<t}\Big)\in\mathbb{R}^{D}.
\end{equation}

\begin{figure}[htbp]
    \centering
    \includegraphics[width=\columnwidth]{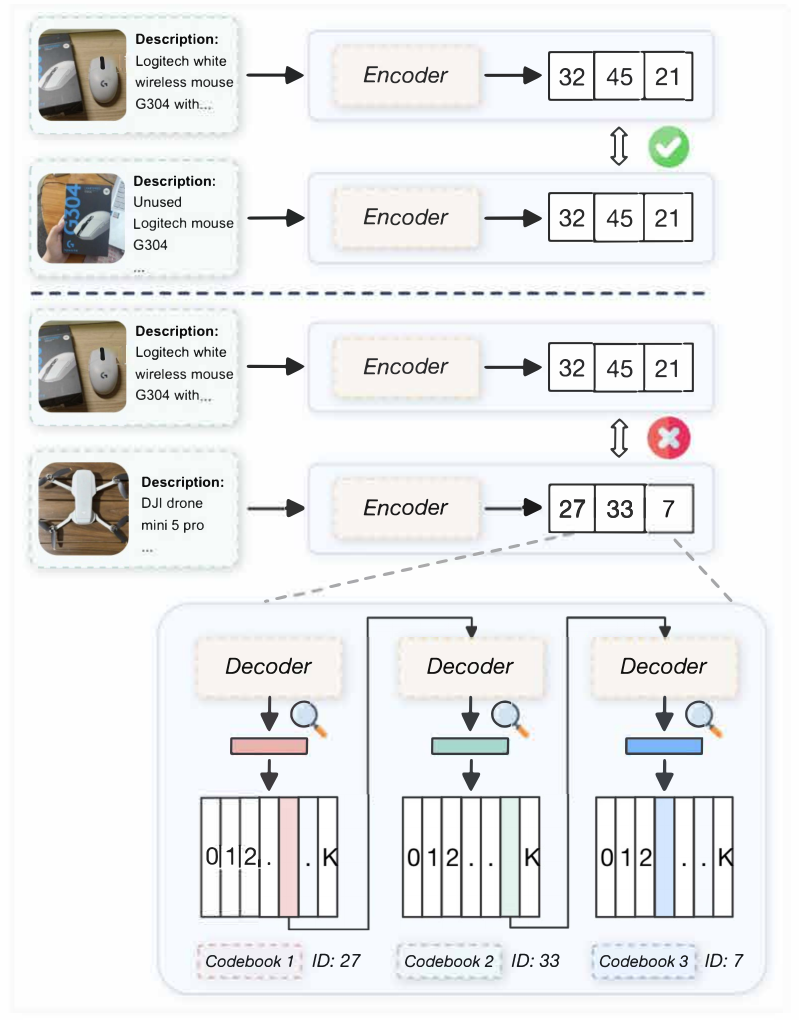}
        \caption{Comparison of Intra- and Inter-Category Semantic IDs.}
    \label{fig:gsid2}
\end{figure}

\paragraph{Hierarchical Code Generation.}
The discrete code $z_t$ is selected via inner-product matching against $E_t$:
\begin{equation}
\small
    Q(z_t\!=\!j\mid z_{<t},b)=\mathrm{softmax}_j\big(d_t \cdot E_t^{\top}\big),~ j\!\in\![K].
\end{equation}
This sequential decoding produces a coarse-to-fine code sequence $(z_1,\dots,z_M)$ as the product's Semantic IDs. As shown in Figure~\ref{fig:gsid2}, semantically proximate items (e.g., two mice) converge at the root index (e.g., 32) and follow an identical refinement path ``32-45-21'', whereas distinct items (e.g., drones) diverge immediately at the root (e.g., 27) to form a separate sequence ``27-33-7''. This contrast between intra- and inter-category cases visualizes how the codebooks encode intrinsic semantic similarity in a data-driven manner.

\paragraph{Final Product Representation.}
Rather than directly using raw encoder features, we take the decoder hidden state at the last step as the product representation. The codebooks impose a coarse-to-fine semantic prior that routes items with shared prefix codes into a common subspace, so that long-tail products can benefit from the semantic structure learned over popular categories.

\paragraph{Training Objective.}
The model is optimized in two stages. The \textit{domain-adaptive pre-training} stage injects e-commerce semantics into the T5 backbone via three sequence-to-sequence tasks: \textit{query generation}, which predicts a search query from a product description; \textit{item cloze}, which reconstructs a product description from its masked version; and \textit{suffix completion}, which generates the suffix of a product description given its prefix. Each task is trained with a standard negative log-likelihood loss, jointly forming
\begin{equation}
\small
    \mathcal{L}_{\text{pre}}=\mathcal{L}_{\text{qry\_gen}}+\mathcal{L}_{\text{item\_cloze}}+\mathcal{L}_{\text{suffix\_comp}}.
\end{equation}
The \textit{progressive training} stage learns Semantic IDs layer by layer. At step $T$, codes $z_{<T}$ are frozen, and the model is trained on $(q, b)$ pairs from search logs, where $q$ is a search query associated with product $b$. The total loss combines a query--item alignment term and a commitment term, $\mathcal{L}=\mathcal{L}_{\text{align}}+\mathcal{L}_{\text{com}}$:
\begin{equation}
\small
\begin{aligned}
    \mathcal{L}_{\text{align}} =& -\log\frac{\exp(q_T\!\cdot\!d_T)}{\sum_{d^{*}\sim\mathcal{B}}\exp(q_T\!\cdot\!d^{*}_T)}\\
    & + \sum_{t=1}^{T}\mathbb{D}_{\text{KL}}\big(P(z_t|z_{<t},q)\,\|\,P(z_t|z_{<t},b)\big),
\end{aligned}
\end{equation}
\begin{equation}
\small
    \mathcal{L}_{\text{com}} = -\sum_{t=1}^{T}\log Q\big(z_t\mid z_{<t},\mathrm{Encoder}(b)\big).
\end{equation}
Codebooks are updated by exponential moving averages for stability. Hard negatives sharing the same code prefix are sampled within each batch to force fine-grained discrimination among semantically similar items.

\paragraph{Three-Stage Multimodal Training.}
So far we have only considered text inputs. To bring images into the same semantic space, we add a separate vision encoder in front of the T5 backbone. We then train the model in three stages. Stage~1 trains the T5 backbone on text alone to establish its textual semantic space. Stage~2 trains a ViT-based vision encoder to align image features with the textual semantic space, while keeping the T5 encoder frozen. Stage~3 then jointly fine-tunes the full model on text, image, and multimodal inputs, producing the final multimodal Semantic IDs. This staged design decouples modalities early to avoid unstable gradient interactions, while enabling efficient end-to-end inference fully compatible with the T5 architecture.

\subsection{Prompt}
\label{appendix_prompt}
In this section, we provide the prompt templates.
\paragraph{Bidirectional Reasoning for Training Data Construction}
In this section, we present the task templates for backward reasoning and forward reasoning in training data construction, as detailed below.

\begin{figure*}[htbp]
    \begin{promptbox}{Prompt for Backward Reasoning}
    \label{prompt:backward_reasoning}
    You are an expert in second-hand product analysis. Your task is to use the information and ground-truth price of Product A to identify strictly comparable products from the retrieved product set that provide a sound basis for pricing.     \\

    \textbf{[Product A]: }\texttt{\{Description of Product A.\}}\\
    
    \textbf{[Price of Product A]: } \texttt{\{Price of Product A.\}}\\

    \textbf{[Retrieved Products]: } \texttt{\{Details of Products B1-B50.\}}\\
    \end{promptbox}
\end{figure*}

\begin{figure*}[htbp]
    \begin{promptbox}{Prompt for Forward Reasoning}
    \label{prompt:forward_reasoning}
    You are an expert in second-hand product pricing. Your task is to use the information about Product A, the complete retrieved product set, and the strictly comparable product subset identified during backward reasoning to construct a coherent and evidence-grounded pricing rationale.\\
    \\
    The price of Product A is provided only as a reference to help align the rationale with the target price. Do not treat this price as a known fact or state that it is provided as input. Instead, derive the rationale naturally from the key attributes of Product A and the retrieved products. Do not work backward from the target price or construct post hoc justifications for it. Every claim, comparison, and pricing consideration in the rationale must be supported by the available product information.\\
        \\
    \textbf{[Product A]: }\texttt{\{Description of Product A.\}}\\
    \\
    \textbf{[Price of Product A --- For Alignment Only]: }
    \texttt{\{Price of Product A.\}}\\
    \\
    \textbf{[Retrieved Products]: } \texttt{\{Details of Products B1-B50.\}}\\
    \\
    \textbf{[Comparable Product Subset]: }
    \texttt{\{Product IDs selected by the backward teacher.\}}\\
                \end{promptbox}
\end{figure*}

\paragraph{Product Retrieval for Price Prediction}
Given a target product and a set of retrieved products with their prices, we prompt the LLM to predict the price of the target product. The full prompt consists of a system prompt and a user prompt, as shown below.
\begin{figure*}[htbp]
    \begin{promptbox}{System Prompt for Price Prediction}
    \label{prompt:system_instruction_reasoning}
    You are an expert in predicting the prices of second-hand products.
    Please provide your pricing rationale and the final predicted price. Wrap the rationale in <reason></reason> and the price in <price></price>.
    
    \end{promptbox}
\end{figure*}

\begin{figure*}[htbp]
    \begin{promptbox}{User Prompt for Price Prediction}
    \label{prompt:second_hand_pricing}
    Please generate a price suggestion for Product A based on the retrieved products (set B). Carefully assess the retrieved products, as not all of them are similar to Product A.
    Your final price suggestion should be derived from a careful analysis of primary pricing factors, such as brand, model, condition, specifications, version and ...
    \\

    \textbf{[Product A]}
    \\
    \texttt{iPhone 16 Pro, White, 256GB, Nearly new, Fully functional.}
    \\

    \textbf{[Retrieved Products]}
    \begin{itemize}[leftmargin=*, itemsep=4pt, topsep=5pt, parsep=0pt, partopsep=0pt]
    \item \texttt{Product B1: Apple iPhone 16 Pro, 256GB, Black, Excellent Condition (90\% Battery). Price: 6088.}
    \item \texttt{Product B2: Apple iPhone 16 Pro, 256GB, White, With a cracked screen. Price: 4999.}
    \item \texttt{...}
    \item \texttt{Product B50: Apple iPhone 15 Pro, 512GB, White, Purchased 1 year ago but barely used (kept idle). In perfect condition. Free charger included. Price: 3999.\\}
    \end{itemize}
    \end{promptbox}
\end{figure*}

\end{document}